\documentclass{article}

\usepackage{PRIMEarxiv}

\usepackage[utf8]{inputenc} 
\usepackage[T1]{fontenc}    
\usepackage{hyperref}       
\usepackage{url}            
\usepackage{booktabs}       
\usepackage{amsmath}        
\usepackage{amsfonts}       
\usepackage{microtype}      
\usepackage{fancyhdr}       
\usepackage{graphicx}       
\usepackage{caption}        
\usepackage{enumitem}       
\setlist[itemize]{leftmargin=*, itemsep=2pt, topsep=4pt}
\usepackage{xcolor}         
\usepackage{tikz}
\definecolor{gsblue}{HTML}{4285F4} 

\DeclareRobustCommand{\gsicon}{%
	\begin{tikzpicture}[baseline=-0.35em]
	\draw[gsblue, fill=gsblue] (0,0) circle [radius=0.16];
	\node[white, font=\bfseries\scriptsize] at (0,0) {G};
	\end{tikzpicture}%
}

\newcommand{\scholarauthorA}{vU6oBhwAAAAJ}  
\newcommand{\scholarauthorB}{rn7gJxMAAAAJ}  
\newcommand{\scholarauthorC}{v5t3VSQAAAAJ}  
\newcommand{\scholarA}{\hspace{0.25em}\href{https://scholar.google.com/citations?user=\scholarauthorA}{\gsicon}}
\newcommand{\scholarB}{\hspace{0.25em}\href{https://scholar.google.com/citations?user=\scholarauthorB}{\gsicon}}
\newcommand{\scholarC}{\hspace{0.25em}\href{https://scholar.google.com/citations?user=\scholarauthorC}{\gsicon}}
\usepackage[round]{natbib}  
\graphicspath{{figures/}}

\hypersetup{
    colorlinks=true,
    linkcolor=black,
    urlcolor=blue,
    citecolor=black,
    pdfborder={0 0 0}
}

\title{AI4SE and SE4AI Exploration: A Decade Looking Back and Forward
\thanks{Preprint. The AI4SE/SE4AI Explorer web application accompanying this article is deployed at \url{https://bankh.github.io/ai4se-se4ai-explorer/} (source code: \url{https://github.com/bankh/ai4se-se4ai-explorer}), and the human--AI agreement dataset is available at \url{https://doi.org/10.7910/DVN/IKLUYN}.}
}

\author{
  Sinan Bank\scholarA\thanks{Corresponding author: \texttt{sinan.bank@colostate.edu}} \\
  Department of Systems Engineering \\
  Colorado State University \\
  Fort Collins, CO 80523 \\
  \And
  Daniel Herber\scholarB \\
  Department of Systems Engineering \\
  Colorado State University \\
  Fort Collins, CO 80523 \\
  \And
  Thomas Bradley\scholarC \\
  Department of Systems Engineering \\
  Colorado State University \\
  Fort Collins, CO 80523 \\
}

\begin{document}
\maketitle

\begin{abstract}
The March 2020 INCOSE INSIGHT special issue on AI and Systems Engineering (SE) became the most downloaded issue in the publication's history and launched a research community that now draws over 250 registrants to its annual workshop. In this article, we trace the progress in AI and SE across three phases---labelled here foundational, applied, and LLM inflection---based on the authors' reading of the field's core papers, and describe our opinions of where the community has converged and where critical gaps remain. Separately, a human--AI agreement literature review leveraging both human expertise and six AI models was performed to assess the relevance of 1,712 INCOSE INSIGHT articles and 889 SERC publications. The results identify five critical research gaps and offer guidance for practitioners navigating AI adoption, assurance, and workforce transformation in SE. We share the agreement data and the AI4SE/SE4AI Explorer web application so readers can compare their own relevance judgments with the human and AI raters.
\end{abstract}


\section{Introduction}

Artificial intelligence (AI) is reshaping how engineers conceive, design, and govern complex systems. In 2020, a landmark INCOSE INSIGHT special issue introduced the AI4SE/SE4AI dual framework---AI to improve systems engineering, and systems engineering for AI-enabled systems---establishing the vocabulary and research agenda for an emerging community \citep{mcdermott2020roadmap}. That issue went on to become the most downloaded in INSIGHT history. Six years later, this research asks the questions of what has advanced, what has surprised, and what steps can practitioners take to shape this transformation?

We instigated this research to assess whether the field's high momentum and activity matches its evidence base. The short answer: not yet. Workshop registration has exceeded 200 every year since the inaugural 2020 event, reaching a sold-out 250-plus in 2025 \citep{serc2025}. The INCOSE International Symposium now hosts a dedicated track on large language models for systems engineering. Yet a recent systematic review identified 284 papers at the AI+SE intersection (especially AI4SE), selecting 33 for in-depth review \citep{poulsen2025}. The field of AI in SE remains empirically nascent.

This narrative draws on two kinds of evidence (both through early 2025). First, we can reference the published literature. This study analyzed 1,712 Google Scholar--indexed INCOSE INSIGHT articles using a human expert and six AI models to identify relevant work at the intersection of AI and systems engineering. Second, we can reference community sources. This study references a corpus of workshop reports, symposium programs, and the SERC publication archive of 889 publications, of which 140 were judged relevant to the topic of AI in SE.

These findings make three contributions to our understanding of the state of the field. First, we trace the trajectory of this field across three proposed phases and seek to describe what aspects of the field the community generally agrees on. Second, we identify five critical research gaps that should be addressed in future studies and symposia. Third, we provide a systematic literature review in the form of a human-model agreement study, share the underlying data, and provide an interactive web application for readers to assess the field using their own judgments.

\section{Methods for a Human-Model Agreement Literature Review}

How do we know which papers belong in this synthesis? Rather than relying on assertion, we designed a study that proved as revealing about AI capabilities as it was about the literature itself.

\noindent\textit{Design.} We asked a human expert and six AI models to independently judge each of 1,712 Google Scholar--indexed INCOSE INSIGHT articles for relevance to the AI--systems engineering domain. The human expert identified 46 articles as relevant. The six models spanned two categories: three proprietary cloud-hosted models (Gemini 3.1 Pro Preview, Claude Opus 4.6, and GPT-5.2) and three locally hosted open-source models (Mixtral 8x22B, Llama 3.3 70B, and DeepSeek R1 70B, via Ollama). For the INSIGHT corpus, all raters---human and AI---judged each article from its title only; for the SERC corpus (889 publications), we ran two input conditions: title-only and title+abstract. We measured each model's alignment with the human expert and its change of mind when given the abstract (transition from title-only to title+abstract judgment). Throughout, the human expert serves as a comparator rather than a gold standard: the agreement statistics measure the concordance of triage-level relevance judgments, not model correctness.

\noindent\textit{What we found.} The results were more nuanced than we expected. We had hypothesized that domain-specific relevance judgment would be difficult for all models, but the proprietary models handled it respectably and consistently, while the local models varied widely. For INSIGHT, percentage agreement with the human ranged from 97.8\% (GPT-5.2) to 99.0\% (Opus 4.6) among proprietary models and from 80.8\% (Mixtral) to 96.7\% (DeepSeek) among local models. For SERC title-only, proprietary models achieved 91.7--96.0\% agreement versus 75.3--91.6\% for local models. Cohen's kappa ($\kappa$), which corrects for chance agreement, sharpens the picture: proprietary models scored $\kappa = 0.67$--$0.77$ (INSIGHT) and $\kappa = 0.66$--$0.85$ (SERC title-only), while local models ranged from near-chance performance (Mixtral, $\kappa = 0.10$ on INSIGHT) to agreement comparable to proprietary models (DeepSeek, $\kappa = 0.71$ on SERC title-only). Put differently, local models disagreed with the human on up to 19.2\% of INSIGHT and 24.8\% of SERC articles, compared with at most 2.2\% and 8.3\% for proprietary models. The key distinction is consistency: proprietary models performed reliably across both corpora, while local model performance was highly model-dependent. The complete agreement statistics and model-by-model breakdowns are available in the provided dataset \citep{bank2025data}.

\noindent\textit{What this means for practitioners.} If AI models disagree this much on what constitutes a ``relevant paper,'' practitioners should expect similar variability when AI tools classify requirements, assess risks, or generate test cases. This is not an argument against using AI tools---it is an argument for understanding their reliability boundaries.

\begin{figure}[!t]
    \centering
    \includegraphics[width=\textwidth]{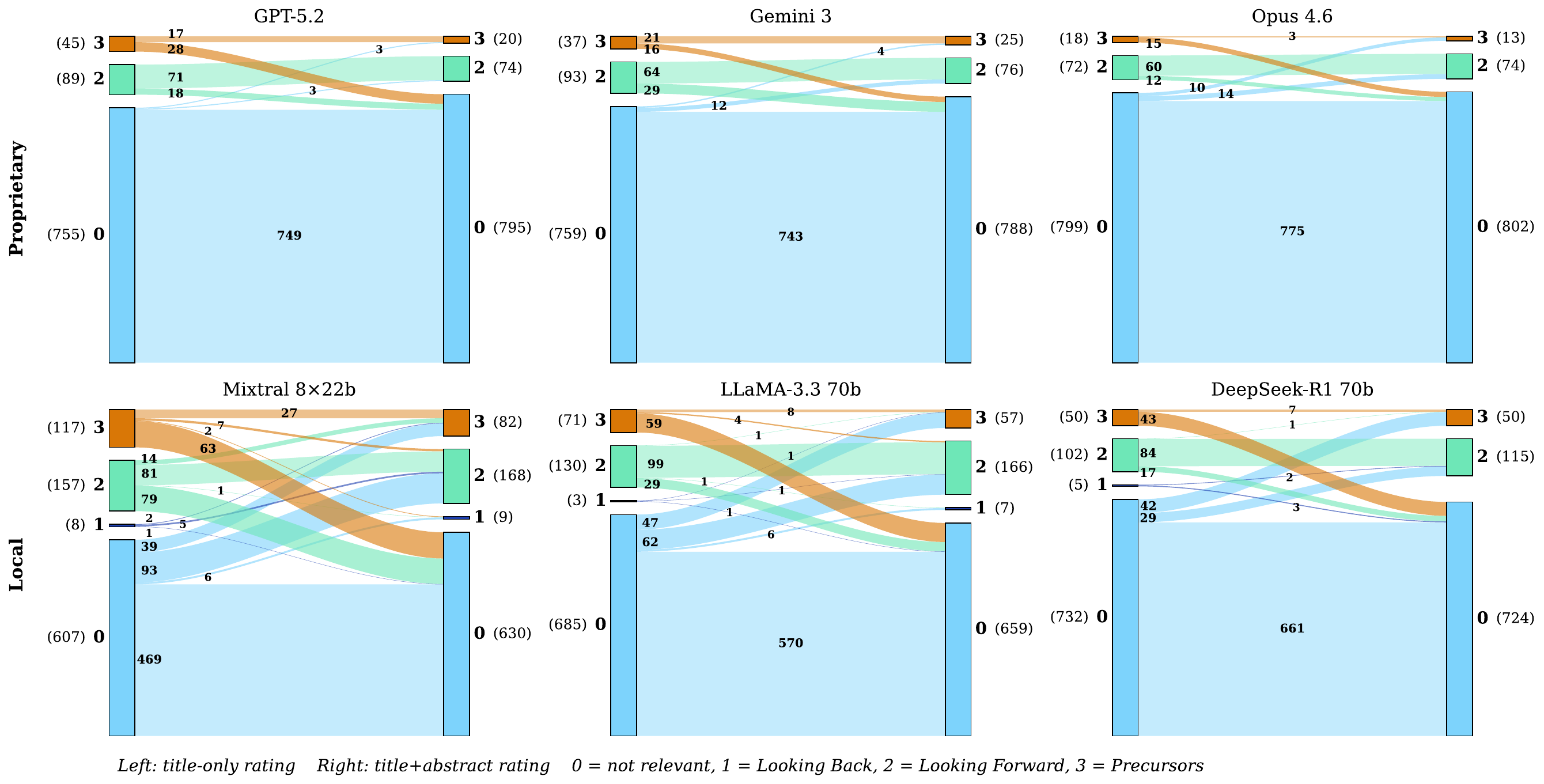}
    \caption{Decision changes of the six models when given the abstract (SERC publications): title-only (left) to title+abstract (right).}
    \label{fig:sankey}
\end{figure}

\noindent\textit{Title-only vs.~title+abstract.} To classify the SERC publications, we compared model judgments under two input conditions: title-only and title+abstract. Under both conditions, each model assigned every publication to one of four categories---not relevant, or relevant as Looking Back, Looking Forward, or Precursors---and we measured how often each model agreed with the human and how often it changed category when the abstract was provided. Figure~\ref{fig:sankey} visualizes these decision changes for all six models as Sankey diagrams, with the width of each link proportional to the number of papers moving from a title-only category (left) to a title+abstract category (right). This view makes the direction of mind change easy to see---for example, how many papers a model kept in the not-relevant category after reading the abstract, or how many it moved from one relevance category to another. Local and proprietary models show transitions from title-only to title+abstract when given the abstract.

\begin{figure}[!b]
    \centering
    \includegraphics[width=0.9\textwidth,height=0.215\textheight,keepaspectratio]{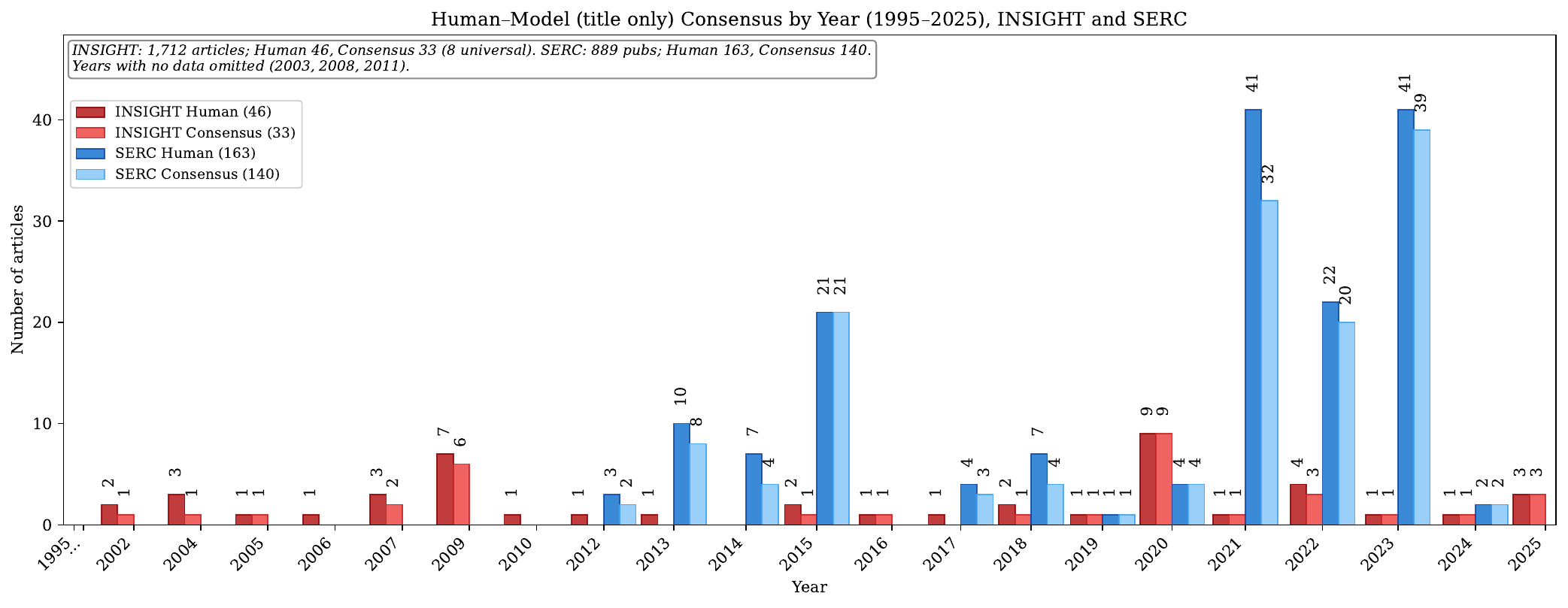}\\[4pt]
    \includegraphics[width=0.9\textwidth,height=0.215\textheight,keepaspectratio]{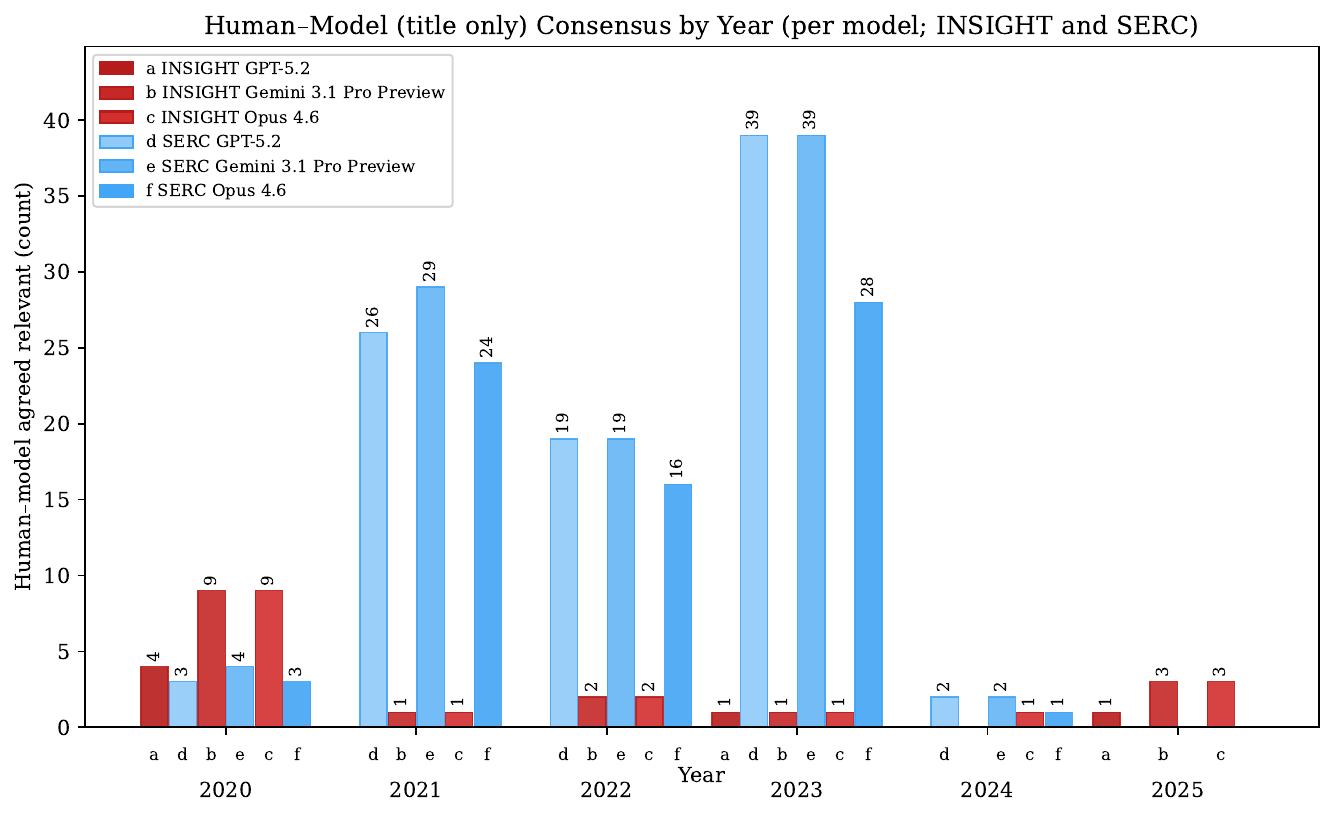}
    \caption{Top: Human--AI agreement by year (1995--2025). Bottom: human--model consensus by year (2020--2025), shown per model, with subpanels a--c for INSIGHT and d--f for SERC. Both panels cover INSIGHT and SERC.}
    \label{fig:agreement}
\end{figure}

\noindent\textit{Year-by-year pattern.} The human expert labeled 46 of the 1,712 INSIGHT articles as relevant. The breakdown by year tells a clear story: before 2020, under 2 percent of INCOSE INSIGHT articles addressed the AI--systems engineering intersection (27 of 1,471); from 2020 onward, that share rose to about 8 percent (19 of 241)---a nearly fourfold increase coinciding with the special issue and subsequent community growth.

Thirty-three articles earned agreement from the human and at least one proprietary AI model (15 from before 2020 and 18 in 2020--2025); eight had universal agreement across all four raters (human and the three proprietary models). Those 33 agreed-relevant articles shape our literature selection and offer a data-backed view of what the community considers core. Figure~\ref{fig:agreement} presents this in two panels. The first (\textit{Human--AI agreement}) shows the full timeline (1995--2025): for each year, how many articles the human labeled relevant (46 total for INSIGHT, 164 for SERC) and how many earned consensus with at least one proprietary model (33 for INSIGHT, 140 for SERC). The second panel (\textit{Human--model consensus}) zooms in on 2020--2025 with one bar per model---subpanels a--c show INSIGHT and d--f show SERC, for GPT-5.2, Gemini 3.1 Pro Preview, and Opus 4.6, respectively. The same article can appear in more than one bar, so the 33 total is not the sum of the bars.

\begin{figure}[!t]
    \centering
    \includegraphics[width=0.45\textwidth]{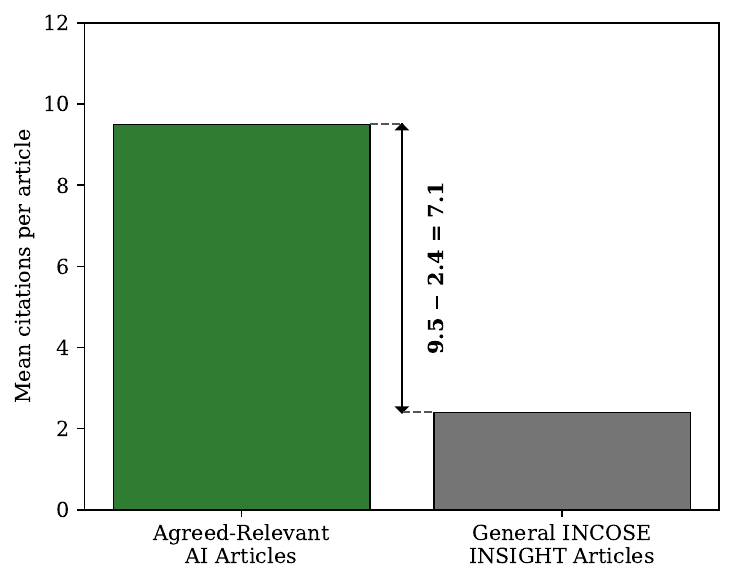}
    \caption{Mean citations per article: agreed-relevant AI articles versus general INCOSE INSIGHT articles.}
    \label{fig:citations}
\end{figure}

\noindent\textit{Citation impact.} Google Scholar citation data (as of 7 February 2026) confirms the impact of these articles. The 33 agreed-relevant AI papers average 9.5 citations per article, compared with 2.4 for the rest of INCOSE INSIGHT---a difference of 7.1 (Figure~\ref{fig:citations}). About 39 percent of the agreed-relevant articles have zero citations; most are newer (post-2021), so citation counts are still accumulating. The most cited are from the 2020 special issue: McDermott et al.'s roadmap leads with 67 citations, followed by Hagedorn et al.\ (40), Freeman (38), Madni (33), and Rouse (28).

\begin{figure}[!b]
    \centering
    \includegraphics[width=0.7\textwidth]{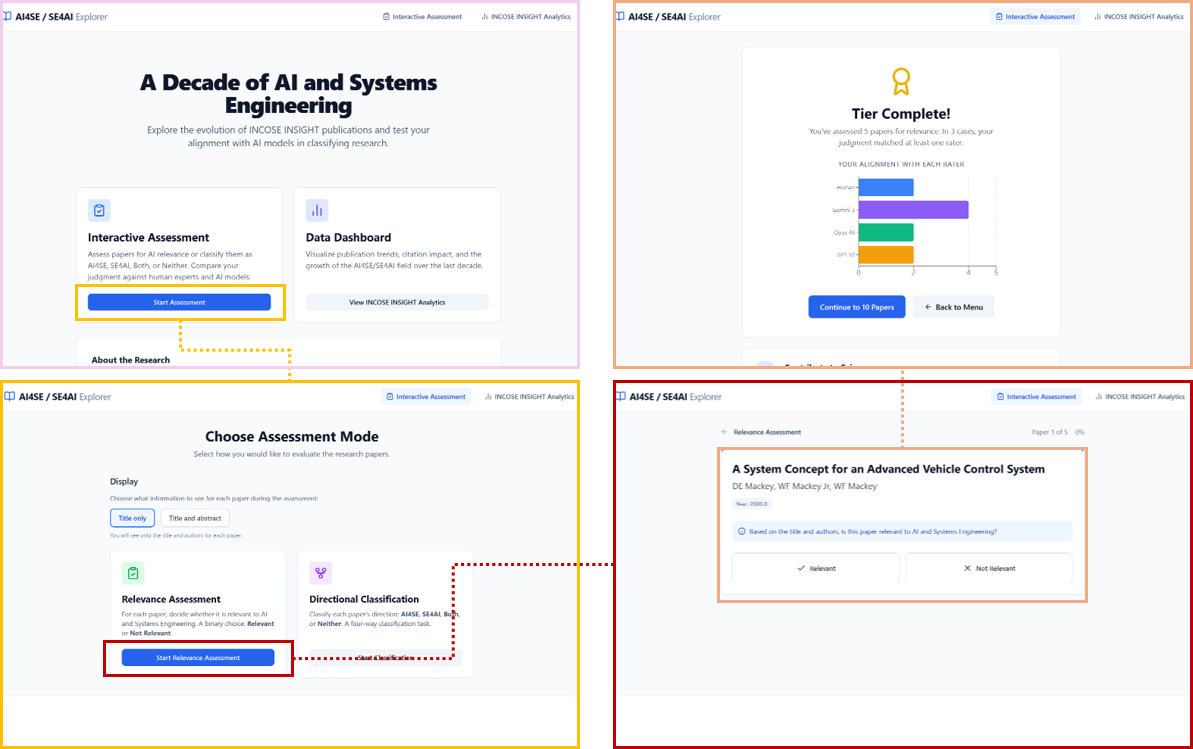}
    \caption{AI4SE/SE4AI Explorer interactive web application for readers to test their relevance judgments against the human and AI raters.}
    \label{fig:explorer}
\end{figure}

To support transparency, we share the agreement data and provide the AI4SE/SE4AI Explorer---an interactive web application where readers can classify INCOSE INSIGHT and SERC publications and compare their judgments with those of the human and AI raters used in our study. Users can filter by corpus (INSIGHT or SERC) and year range, then assess papers for relevance or classify them by direction (AI4SE, SE4AI, Both, or Neither). After each relevance assessment, the application reveals how the human expert and all six AI models rated that paper. Users can submit their evaluation results back to the public repository, contributing to a growing dataset of human-AI agreement beyond the original study. The Explorer also includes an interactive dashboard that displays the agreement data by year, per-model consensus breakdowns, and publication analytics (Figure~\ref{fig:explorer}). All source code is available online \citep{bank2025explorer}, and the agreement data is archived at Harvard Dataverse \citep{bank2025data}.

\noindent\textit{Limitations.} Four limitations bound these results. First, the human reference judgments come from a single expert rater without an intra-rater reliability check; the agreement statistics therefore measure concordance with one informed reader, not with a consensus ground truth. Second, screening at the title and abstract level rests on the assumption that titles and, where used, abstracts faithfully signal a paper's content. This assumption is weakest for the INSIGHT corpus, a practitioner magazine we screened on titles alone, so its relevance judgments depend entirely on how well titles capture content. For SERC we could test the assumption directly: the two-condition comparison (title-only versus title+abstract) is our empirical check---adding the abstracts left the proprietary models' judgments largely stable (about 6\% of papers reclassified), while the local models proved far more sensitive (15--37\% of papers reclassified)---so for the reliable proprietary models the abstract rarely overturned a title-level judgment, whereas the local models' volatility is a further sign of their instability on this task. Third, Cohen's $\kappa$ is sensitive to category prevalence, and relevant-article prevalence differs substantially between the INSIGHT ($\approx$3\%) and SERC ($\approx$16--18\%) corpora, so $\kappa$ values should be compared across corpora with caution. Fourth, the citation-impact comparison is confounded by venue visibility and topic recency: the most-cited agreed-relevant articles appear in a highly downloaded special issue on a fast-growing topic.

\section{The First Six Years in Three Phases}

Based on the authors' observations, this paper asserts that the period 2020--2025 can be divided into three recognizable phases, each building on the last and moving from conceptual architecture to applied experimentation, and finally to the large language model inflection. These phases are not sequential replacements but cumulative layers---foundational work continues as applied, and LLM-driven research builds upon it. Figure~\ref{fig:timeline} illustrates the trajectory of relevant publications in INCOSE INSIGHT and in SERC reporting. Our three proposed phases of activity are as follows:

\begin{figure}[!htb]
    \centering
    \includegraphics[width=0.85\textwidth]{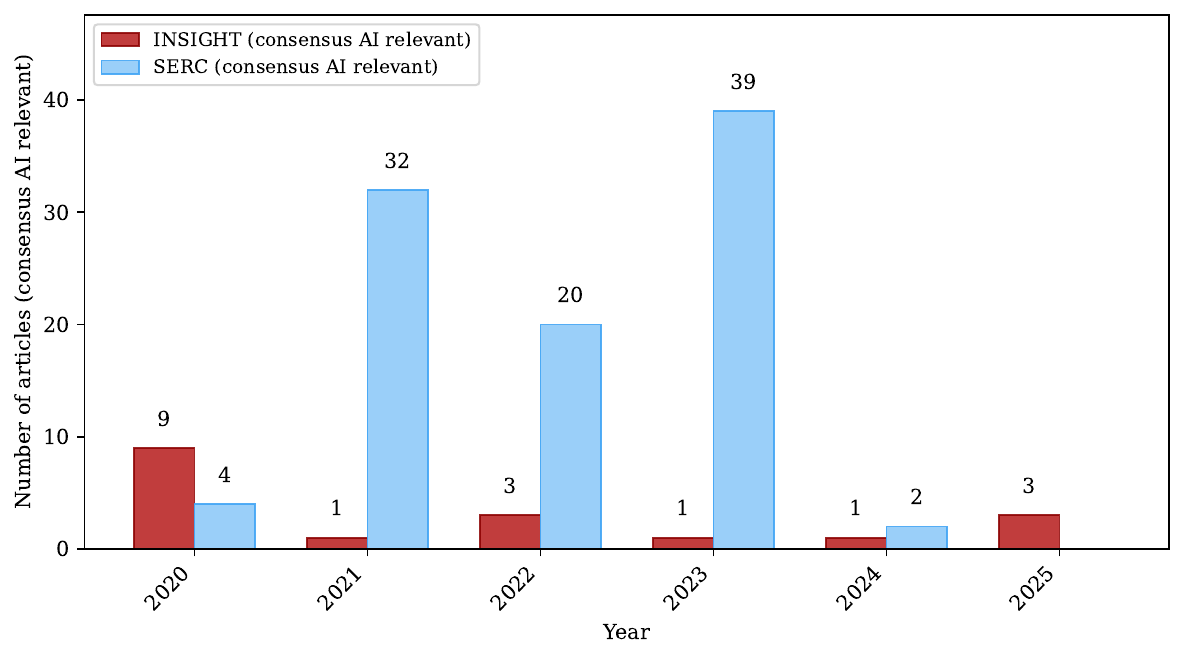}
    \caption{Consensus AI-relevant publication count by year (2020 onwards): INSIGHT (red) and SERC (light blue) \citep{serc2025}.}
    \label{fig:timeline}
\end{figure}

\noindent\textit{Phase 1: Foundational (emerging 2020--2021).} The first phase established many of the conceptual definitions of AI in SE that the SE community still uses. \citet{mcdermott2020roadmap} introduced the AI4SE/SE4AI research roadmap and won the INCOSE 2020 Best Article Award. \citet{madni2020} reframed AI as augmented intelligence---augmentation, not replacement---which has become a strong community understanding of the role of AI in SE processes. \citet{rouse2020} articulated a perspective labelled ``AI as systems engineering,'' envisioning AI cognitive assistants that understand and guide systems engineering concepts and workflows. \citet{freeman2020} identified key themes for how test and evaluation must evolve for AI-enabled systems. \citet{hagedorn2020} demonstrated the use of ontology-based knowledge representation to promote augmented and artificial intelligence in systems engineering. These works have largely set the vocabulary and research agenda that subsequent phases build upon.

\noindent\textit{Phase 2: Applied (emerging 2022--2023).} As the field moved from theory and definitions to practice, several research themes matured in parallel:

\begin{itemize}
    \item NLP for requirements management emerged as a dominant theme and key application in the INCOSE International Symposium proceedings \citep{riesener2021, kulcsar2022}.
    \item Ontology-based knowledge graphs have matured into integration frameworks for digital engineering \citep{dunbar2023}.
    \item Safety assurance for ML systems has become a focus, with researchers addressing certification challenges for AI-based airborne systems \citep{paramasivam2023} and safety assurance of autonomous systems incorporating ML using model-based approaches \citep{zeller2023}.
    \item Roadmap updates: \citet{pepe2022} articulated the long-term human-machine co-learning vision, drawing on four SERC and US Army workshops to define a path from current AI-assisted tools toward genuine human-machine partnership.
\end{itemize}

In this period, the AI4SE and SE4AI workshop---which has drawn over 200 registrants annually since its inaugural virtual event in 2020---added dedicated tracks on trustworthy AI and human-AI teaming. The SERC publication archive reflects this surge: 39 human--AI agreed-relevant papers in 2023 alone (of 889 SERC publications total, 140 agreed-relevant across all years), covering topics from generative AI for MBSE to cost-aware Bayesian agents for human-AI teaming \citep{serc2025}.

\noindent\textit{Phase 3: LLM Inflection (emerging 2024--2025).} The authors assert here that the transition into a new phase in the last 1--2 years has been genuinely surprising. We expected steady, incremental progress, but what we observed was an inflection that neither the 2020-era roadmaps nor perhaps many researchers had fully anticipated. The LLM Inflection phase is characterized as a phase where AI for SE has developed rapidly into a value-driven, validation-focused approach.

Consider a concrete example. A systems engineer sits down with AI-SME, a CATIA Magic plugin powered by GPT-4 Turbo for model-based systems engineering \citep{johns2024}, and types: ``Create a SysML block definition diagram for a thermal management subsystem with liquid cooling, heat exchangers, and fault monitoring.'' Within seconds, the tool generates a structured model---blocks, ports, flow connections---that would have taken an experienced modeler an hour to draft from scratch. The skeleton is genuinely useful. But the generated model also omits a critical thermal-structural interface, assigns a flow port where a standard port is needed, and names a block in a way that violates the project's naming conventions. The systems engineer spends twenty minutes refining rather than sixty minutes creating. That interaction---useful draft, critical errors, human refinement and validation required---captures where the field stands today.

Beyond AI-SME, researchers have shown that LLMs can generate SysML v2 models directly from natural language \citep{dehart2024}. Multi-agent architectures such as MITRE's SELMA framework achieved strong performance on architecture queries \citep{mitre2024selma}. \citet{mcdermott2024roadmap} published the updated SERC AI and Autonomy Roadmap, integrating generative AI. Risk assessment frameworks for AI applications emerged to address this new class of tools.

By 2025, workshop registration exceeded 250, and the event sold out, with keynotes from OpenAI and the Chief Digital and AI Office. Demographics from the 2025 workshop---roughly 39\% academia, 38\% industry, 16\% government, and 7\% FFRDCs---show that the field has moved well beyond a narrow research community \citep{serc2025}.

We flag Phase~3 as the most interpretive of the three: by human--AI agreed-relevant publication count it is also the thinnest, with fewer agreed-relevant articles in 2024--2025 than in the foundational and applied phases (Figure~\ref{fig:timeline}). This dip is partly an artifact---our data extend only through early 2025, so 2025 is incomplete, and very recent work has not yet been fully indexed or cited. Accordingly, the inflection registers less as a jump in publication volume than as a steep and accelerating rise in interest in the domain: workshop attendance climbing every year to a sold-out 2025 event with headline keynotes, a dedicated INCOSE Symposium track on large language models for systems engineering, the landmark tool demonstrations above, and an audience that now reaches well beyond academia into industry and government. Archival publication output typically trails such a surge of interest by one to two years, so for this phase the interest signal, more than the publication count, marks the inflection. Table~\ref{tab:phases} summarizes the three phases.

\begin{table}[!htb]
    \centering
    \caption{The First Six Years in Three Phases (2020--2025).}
    \label{tab:phases}
    \begin{tabular}{@{}p{0.155\linewidth} p{0.095\linewidth} p{0.155\linewidth} p{0.36\linewidth} p{0.115\linewidth}@{}}
        \toprule
        \textbf{Phase} & \textbf{Period} & \textbf{Theme} & \textbf{Key Markers} & \textbf{Primary Challenge} \\
        \midrule
        1. Foundational & 2020--2021 & Conceptual architecture & McDermott AI4SE/SE4AI roadmap, Madni augmented intelligence, Rouse AI as SE, Freeman T\&E framing & Definition \\
        \addlinespace
        2. Applied & 2022--2023 & Experimentation & NLP for requirements, ontology-based knowledge graphs, safety assurance for ML systems, workshop growth to 200+ & Application \\
        \addlinespace
        3. LLM Inflection & 2024--2025 & Acceleration & GPT-4 in MBSE (AI-SME, SELMA), LLM-based SysML v2, multi-agent architectures, risk frameworks for LLMs & Validation \\
        \bottomrule
    \end{tabular}
\end{table}

\section{Where the Community Converges}

Having traced these first years, we can now ask: where does the research and practice community converge on an understanding of the state of the field? Despite rapid evolution, we observe that several themes have achieved a broad consensus.

\noindent\textit{Augmented Intelligence Thesis.} Madni's 2020 reframing of AI as augmentation rather than automation has become the community default. Every subsequent SERC roadmap update, INCOSE working group output, and Vision 2035 scenario reinforces the human-AI teaming paradigm \citep{incose2022vision}.

The empirical evidence supports this. AI tools for model-based systems engineering work best as drafting aids requiring human refinement---as the AI-SME vignette above illustrates. AI-generated system models produce useful drafts but contain critical errors when unsupervised. Human-in-the-loop design is the shared vision.

\noindent\textit{AI4SE/SE4AI Duality.} There is an emerged consensus that both research directions are essential, though they serve different needs. AI4SE---using AI to improve systems engineering---dominates published volume, especially natural language processing and large language models for requirements. SE4AI---systems engineering for AI-enabled systems---carries greater institutional weight, driving standards development and acquisition policy.

Whereas much of the literature described above concentrates on AI4SE, the maturation of the NIST AI Risk Management Framework \citep{tabassi2023nist}, ISO/IEC 42001 \citep{iso2023}, IEEE 7000 \citep{ieee2021}, and DoD Responsible AI frameworks \citep{dod2022} reflects evidence that the SE4AI framework has also become imperative. AI governance is no longer optional; systems engineers operating in regulated or acquisition contexts must map these frameworks to their lifecycle processes. Practitioners need both: tools to augment their work, and methods to engineer trustworthy AI systems.

\section{Critical Gaps and Open Questions}

Based on this understanding of the state of the AI4SE/SE4AI field, the authors have identified five gaps that we believe deserve explicit attention.

\noindent\textit{Empirical Validation Deficit.} Most AI4SE tools have been demonstrated on toy problems or single case studies. Industrial-scale validation is largely absent in the literature. The systematic review by \citet{poulsen2025} identified 284 papers at the AI+SE intersection (especially AI4SE), selecting 33 for in-depth review. The SERC archive (889 publications) includes 140 human--AI agreed-relevant papers, but few are controlled experiments or longitudinal field studies. Until the community invests in rigorous, replicable validation---ideally with industrial partners---AI4SE tools will remain promising demonstrations rather than proven capabilities with industrial and practical validity.

\noindent\textit{Systems Engineering vs. Software Engineering Conflation.} Many ``AI for systems engineering'' papers actually address software engineering tasks: code generation, test automation, or CI/CD pipeline optimization. Systems engineering's unique challenges---lifecycle integration across hardware, software, and human factors; multi-stakeholder coordination; physical-cyber coupling---are distinct from software development concerns. \citet{poulsen2025} found that most candidate studies had to be excluded for precisely this reason. We see this conflation as one of the field's most persistent obstacles.

\noindent\textit{Systems Engineering--Specific Benchmarks.} SysEngBench \citep{bell2024sysengbench} is the only known domain-specific benchmark for assessing AI tools in systems engineering tasks. The software engineering community has developed dozens of benchmarks for code generation, bug detection, and documentation---systems engineering has one. Without shared benchmarks, the community cannot compare tools, track progress, or establish minimum competency thresholds.

\noindent\textit{Traceability for AI-Generated Artifacts.} No mature framework exists for maintaining traceability and provenance when AI generates or modifies systems engineering artifacts such as requirements, SysML models, or architecture documents. For regulated industries---defense, aerospace, medical devices, nuclear---this gap is critical. Certification and audit processes depend on knowing who (or what) produced each artifact and how it was validated. Researchers have begun early-stage work on AI-assisted traceability \citep{bonner2024}, but much remains to be done.

\noindent\textit{Workforce Transformation.} INCOSE Vision 2035 \citep{incose2022vision} describes a transformed practice with AI integration, but concrete curricula, competency models, and transition roadmaps for AI-literate systems engineers remain underdeveloped. As AI tools become embedded in everyday workflows, the question shifts from ``will systems engineers use AI?'' to ``what must systems engineers understand about AI to use it responsibly?''

\section{Implications for Practitioners}

In keeping with this article's practitioner focus, we can also take the opportunity here to direct SE practitioners with guidance organized around what to adopt, what to approach with caution, and what to monitor.

\noindent\textit{Adopt: Where AI Tools Are Ready.} Several AI capabilities have reached sufficient maturity for careful adoption:

\begin{itemize}
    \item NLP for requirements classification and elicitation has been demonstrated across multiple studies and is available in commercial tools, allowing organizations to begin low-risk pilot deployments.
    \item AI-assisted drafting of specifications and system descriptions can accelerate early-phase work, provided a human expert reviews and validates output against the system context.
    \item Ontology-aided knowledge integration supports interoperability in digital engineering across heterogeneous toolchains \citep{dunbar2023}.
    \item LLM-based querying of system models---as demonstrated by AI-SME \citep{johns2024}---shows promise for making model repositories accessible to non-specialist stakeholders.
\end{itemize}

\noindent\textit{Use Caution: Where the Risks Outweigh the Benefits Today.} AI4SE applications, such as autonomous verification and validation without human oversight, safety-critical decision automation, and unsupervised generation of contractual or regulatory artifacts, all warrant caution. Our own agreement study illustrates the variability: even among proprietary AI models, relevance judgments varied substantially, and local open-source models ranged from near-chance to proprietary-comparable agreement depending on the model. The augmented intelligence paradigm---AI as a drafting aid, humans as the final authority---should guide adoption decisions.

\noindent\textit{Invest Now.} Three investments merit immediate attention:

\begin{itemize}
    \item Traceability infrastructure for AI-generated artifacts. Establish provenance tracking before AI tools become deeply embedded in workflows---retrofitting traceability is far more expensive than building it in from the start.
    \item Human-in-the-loop design patterns. Rather than bolting human oversight onto AI tools after deployment, design the human-AI interaction from the start: where does AI propose, where do humans decide, and where does the system learn from feedback?
    \item Systems engineering benchmarks. The field needs shared evaluation frameworks, and organizations that participate in their development will shape the standards that govern the qualification of AI tools.
\end{itemize}

\noindent\textit{Monitor.} Three areas merit watching rather than immediate investment: fully autonomous co-engineering, where AI acts as a peer rather than an assistant; full-lifecycle AI integration from concept through disposal; and emerging certification frameworks for AI-intensive systems. We assert that these are active research areas where both the technology and the governance landscape are still shifting rapidly, but where the potential for the next point of inflection in our field will emerge.

\section{Conclusion}

The first six years of the decade have transformed AI and systems engineering from a conceptual roadmap into a rapidly expanding field of research and practice. Three phases---foundational, applied, and LLM inflection---are proposed here to capture that trajectory. We identify that the community has largely agreed on augmented intelligence as the paradigm, the necessity of both AI4SE and SE4AI, and the inevitability of standards-driven AI governance.

Our human-AI agreement methods as presented here and in the AI4SE/SE4AI Explorer web application provide a set of systematic review data to support the assertions documented in this paper. The methods also invite critical thinking into the validity and capability of AI tools for SE. We identified 33 articles where human and AI reviewers concur on their relevance to the AI in SE domain, and found that proprietary models handled this domain-specific judgment task with moderate to strong reliability, while open-source local models ranged from near-chance agreement to agreement comparable to proprietary models depending on the model. We share the agreement data and the AI4SE/SE4AI Explorer web application to support transparency and invite the community to contribute to future work on human-model agreement in literature curation.

What practitioners can do now is concrete. Critical gaps remain in empirical validation, domain-specific benchmarks, traceability, and workforce transformation. The research and knowledge production community will require practitioners' participation to close these research gaps in the coming years. There is great potential for practitioners to be at the center of the next revolutionary developments in AI4SE and SE4AI. The remainder of the decade will determine whether the community's vision of human-machine co-learning becomes reality. We believe the foundation for value-driven and effective AI is in place for the SE community. Systems engineers must prioritize evidence over enthusiasm.

\section*{Data and Software Availability}
The human--AI agreement dataset is archived at Harvard Dataverse (\url{https://doi.org/10.7910/DVN/IKLUYN}). The AI4SE/SE4AI Explorer web application is deployed at \url{https://bankh.github.io/ai4se-se4ai-explorer/}, and its source code is maintained at \url{https://github.com/bankh/ai4se-se4ai-explorer}.

\bibliographystyle{apalike}
\bibliography{references}

\end{document}